\documentclass[runningheads]{llncs}
\usepackage{graphicx}
\usepackage{amsmath,amssymb}
\usepackage{verbatim}
\usepackage[utf8]{inputenc}
\usepackage[section]{placeins}
\usepackage{float} 
\usepackage{color}
\usepackage{multirow}
\usepackage{algorithmic}
\usepackage{algorithm}
\usepackage{cite}
\usepackage{subfigure}
\usepackage{booktabs}       
\usepackage{url}

\newcommand{\sh}[1]{\textcolor[rgb]{1,0,0}{{#1}}}

\begin{document}
\title{Feature Re-calibration based Multiple Instance Learning for Whole Slide Image Classification}

\titlerunning{Feature Re-calibration based MIL for Whole Slide Image Classification}

\author{Philip Chikontwe \inst{1} \and Soo Jeong Nam \inst{2} \and Heounjeong Go \inst{2,3} \and Meejeong Kim \inst{2} \and Hyun Jung Sung \inst{2} \and Sang Hyun Park \inst{1}}

\authorrunning{Chikontwe et al.}
\institute{******************}

\institute{Department of Robotics and Mechatronics Engineering, DGIST, Korea \and
Department of Pathology, Asan Medical Center \\ \and Department of Pathology, University of Ulsan College of Medicine \\
\email{philipchicco@dgist.ac.kr, shpark13135@dgist.ac.kr} }

\maketitle              

\begin{abstract}
	
Whole slide image (WSI) classification is a fundamental task for the diagnosis and treatment of diseases; but, curation of accurate labels is time-consuming and limits the application of fully-supervised methods. To address this, multiple instance learning (MIL) is a popular method that poses classification as a weakly supervised learning task with slide-level labels only. While current MIL methods apply variants of the attention mechanism to re-weight instance features with stronger models, scant attention is paid to the properties of the data distribution. In this work, we propose to re-calibrate the distribution of a WSI bag (instances) by using the statistics of the max-instance (critical) feature. We assume that in binary MIL, positive bags have larger feature magnitudes than negatives, thus we can enforce the model to maximize the discrepancy between bags with a metric feature loss that models positive bags as out-of-distribution. To achieve this, unlike existing MIL methods that use single-batch training modes, we propose balanced-batch sampling to effectively use the feature loss i.e., (+/-) bags simultaneously. Further, we employ a position encoding module (PEM) to model spatial/morphological information, and perform pooling by multi-head self-attention (PSMA) with a Transformer encoder. Experimental results on existing benchmark datasets show our approach is effective and improves over state-of-the-art MIL methods. \url{https://github.com/PhilipChicco/FRMIL}   

\end{abstract}
\graphicspath{ {./figures/} }
\section{Introduction}
Histopathology image analysis (HIA) is an important task in modern medicine and is the gold standard for cancer detection and treatment planning\cite{he2012histology}. The development of whole slide image (WSI) scanners has enabled the digitization of tissue biopsies into gigapixel images and paved the way for the application of machine learning techniques in the field of digital pathology\cite{banerji2022deep,li2021comprehensive}. However, employing popular convolutional neural  network (CNN) architectures for varied tasks in HIA is non trivial and has several challenges, ranging from the large size of WSIs and extreme high resolution to lack of precise labeling and stain color variations\cite{li2021comprehensive}. This motivates the need for memory efficient methods that mitigate the need for fine-grained labels and are fairly interpretable. To address this, multiple instance learning (MIL)\cite{wang2018revisiting,amores2013multiple} is a popular formulation that considers diagnosis as a weakly supervised learning problem\cite{srinidhi2021deep}. 

Through the recent advances in deep learning\cite{vaswani2017attention,he2016deep}, MIL based histopathology\cite{chen2021pixel,fan2021learning,sharma2021cluster,rymarczyk2021kernel,chikontwe2020multiple} analysis has achieved notable success\cite{dimitriou2019deep,lu2021data,shi2020loss,ilse2018attention}. For instance, Li \textit{et al.}\cite{li2021dual} introduced non-local attention to re-weight instances relative the highest scoring instance (critical) in a bag, proving to be a simple yet effective approach. However, the critical instance is only employed for implicit instance re-weighting and the method is sensitive to both the choice of the instance feature encoder (i.e., pre-trained ImageNet or self-supervised), and the scale of patches used. In MIL-RNN\cite{campanella2019clinical}, recurrent neural networks (RNN) are used to sequentially process instance features, partially encoding position and context, but is limited in the ability to capture long range dependences. 

Thus, follow-up works\cite{lu2021data,li2021dual,shao2021transmil,li2021dt} built on the latter with more complex attention-based variants using Transformer\cite{vaswani2017attention,dosovitskiy2020image} inspired architectures to better model long range instance correlations via multi-head self-attention (MSA) with positional information encoding. Along this line of thought, TransMIL\cite{shao2021transmil} highlights the importance of spatial positional encoding (PE) and single-scale learning over the latter, but is relatively sensitive to the depth of PE layers (i.e., x3) and does not explicitly pool all instances to a single bag representation, instead uses a learnable class token for final bag-level prediction. Thus, the use of Transformers with several MSA blocks can be computationally prohibitive, and would be more desirable to have less over-parameterized designs. 

To address these challenges, we propose a \textbf{F}eature \textbf{R}e-calibration based MIL framework (FRMIL), building upon prior MIL approaches\cite{shao2021transmil,li2021dual,li2021dt} leveraging MSA with Transformer encoders. Here, we argue that re-calibrating the distribution of instance features can improve model performance towards better generalization by using the properties of the data distribution directly. In vision tasks such as few-shot learning, feature/distribution re-calibration is used to enable better generalization when learning from limited samples by transferring statistics from classes with sufficient examples\cite{yang2020free}. However, in the MIL scenario, instances are not always i.i.d \cite{shao2021transmil}, especially since positive instances are often limited in a WSI i.e., ($\le 10\%$). Thus, we consider a simpler form that uses the max instance to shift the original distribution towards better separability. Also, we consider MIL and anomaly detection\cite{chalapathy2019deep,feng2021mist,lee2021weakly} as closely related tasks. For instance, Lee \textit{et al.}\cite{lee2021weakly} leveraged MIL for weakly-supervised action localization by modeling background/normal actions as out-of-distribution using uncertainty by considering their inconsistency in a sequence with video-level labels only. 

\begin{figure}[t!]
	\begin{tabular}{ccc}
		\includegraphics[width=0.32\linewidth]{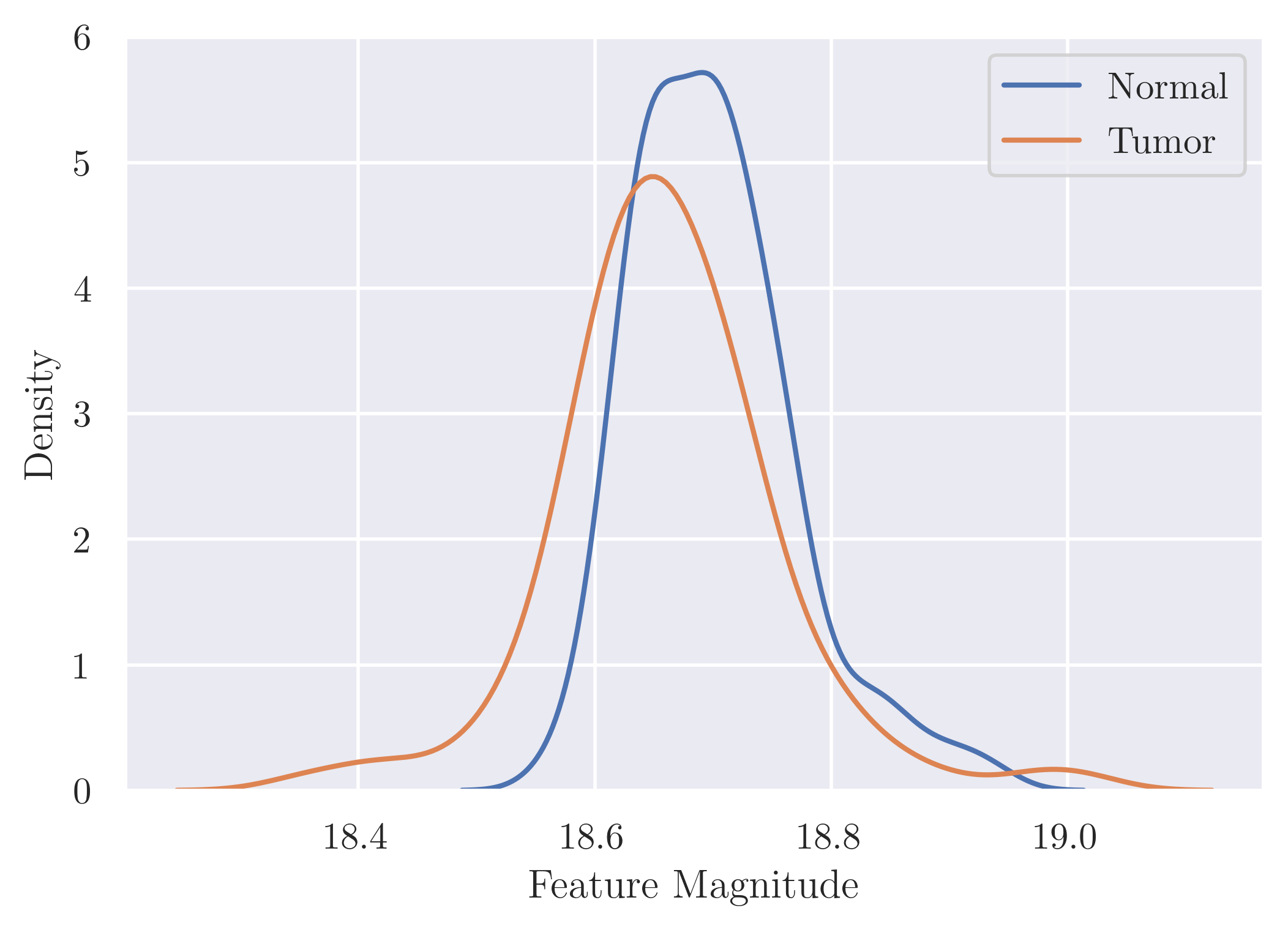} &
		\includegraphics[width=0.32\linewidth]{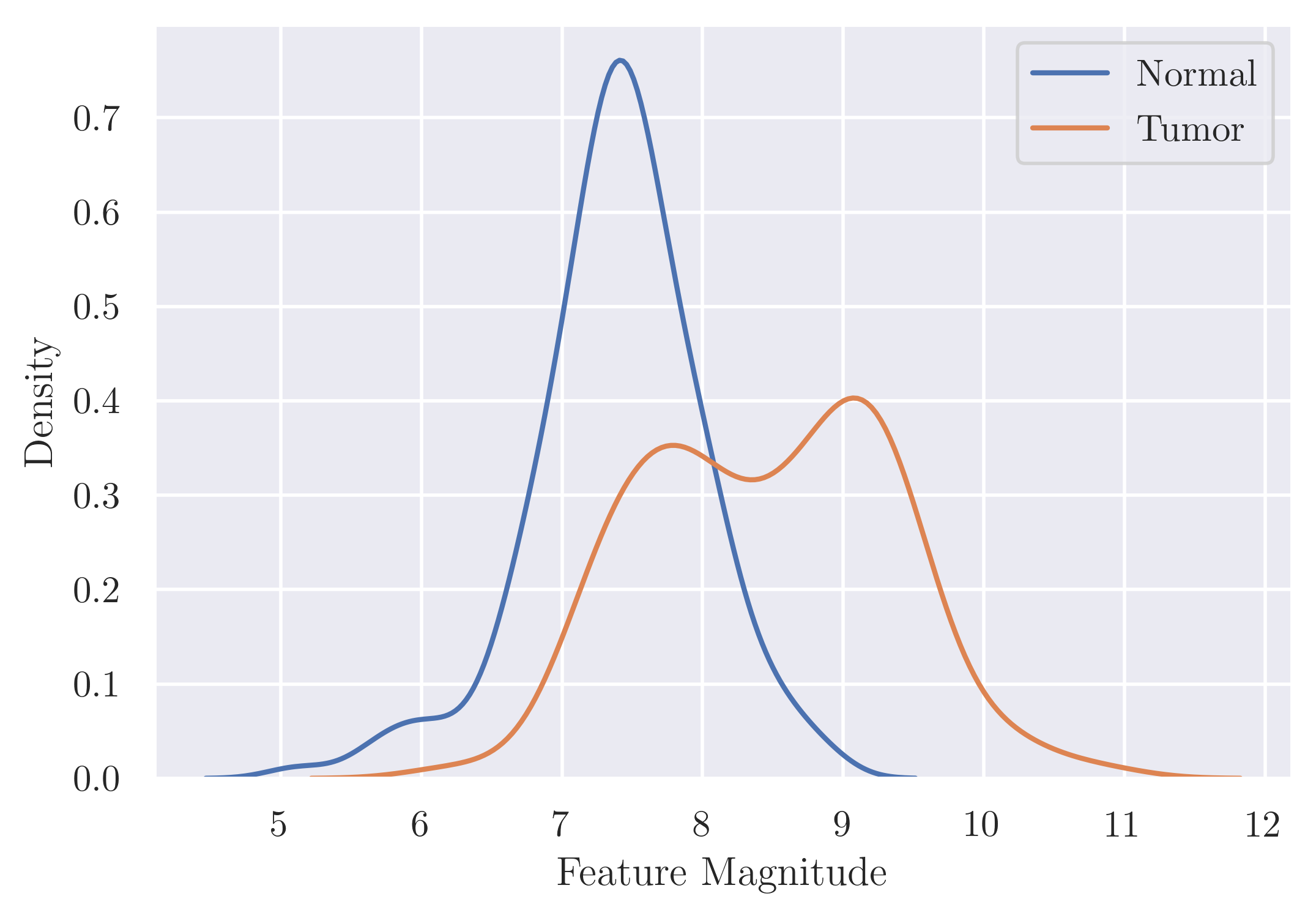} &
		\includegraphics[width=0.32\linewidth]{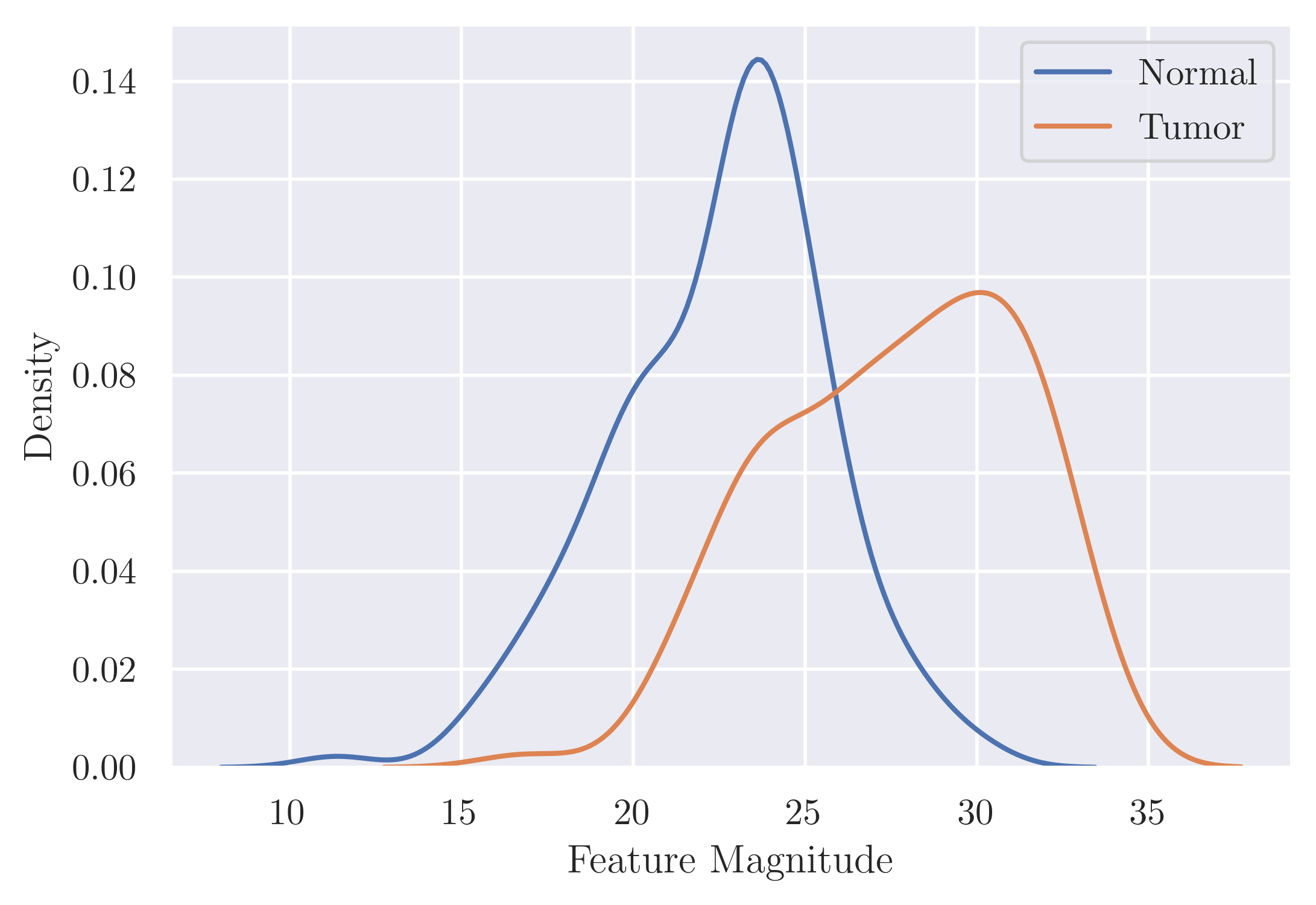} 
		\\
		$44.96\%$ & $83.72\%$ (\sh{$+39$}) & $89.10\%$ (\sh{$+44$})
		\\
		(a) & (b) & (c)
	\end{tabular}
	\caption{Normalized density plots of the mean feature magnitudes on the CAMELYON16\cite{bejnordi2017diagnostic} train-set, with test-set accuracy and improvements (red color). (a) Original feature magnitudes. (b)  Max-instance calibrated based features. (c) Features learned by our FR-MIL model.} \label{fig2}
\end{figure}

Inspired by these works, we hypothesize that features from positive and negative bags (binary MIL) exhibit larger and smaller feature magnitudes respectively, and this prior can be directly encoded into the learning framework for better representation learning\cite{lee2021weakly}. In Fig. \ref{fig2}, we show this phenomena and our intuition to highlight how the standard MIL assumption of having at-least one (+) instance in a bag can be used to make the distribution more separable. Herein, we establish a simple non-parametric baseline that re-calibrates features by subtracting the max instance per-bag, and then computes the probability of a bag-label as the normalized minimum between the mean magnitude and the estimated bag magnitude (see Sec. \ref{sec:methods}). Our evaluation shows that the baseline performance is comparable to classic MIL operators (i.e., max/mean-pooling)\cite{wang2018revisiting}. 

To incorporate this idea in our framework, we explicitly re-calibrate features with the aforementioned concept, and then feed the new features to a positional encoding module (PEM)\cite{shao2021transmil} followed by a single pooling multi-head self-attention block (PMSA)\cite{lee2019set} for bag classification. To effectively enforce feature magnitude discrepancy, we propose a feature embedding loss that maximizes the distance between positive and negative bag features, as well as the standard cross-entropy losses. The main contributions of this work are as follows: (i) We show that feature re-calibration using the max-critical instance embedding is a simple yet powerful technique for MIL, (ii) We introduce a feature magnitude loss to learn better instance/bag separation, (iii) To obtain robust bag embeddings, we leverage a positional encoder and a single self-attention block for instance aggregation, and (iv) Experimental results on a public benchmark and inhouse-curated datasets demonstrate the effectiveness of our method over state-of-the-art methods.

\section{Methods}
\label{sec:methods}
\begin{figure}[t!]
	\centering
	\includegraphics[width=0.85\linewidth,height=5cm]{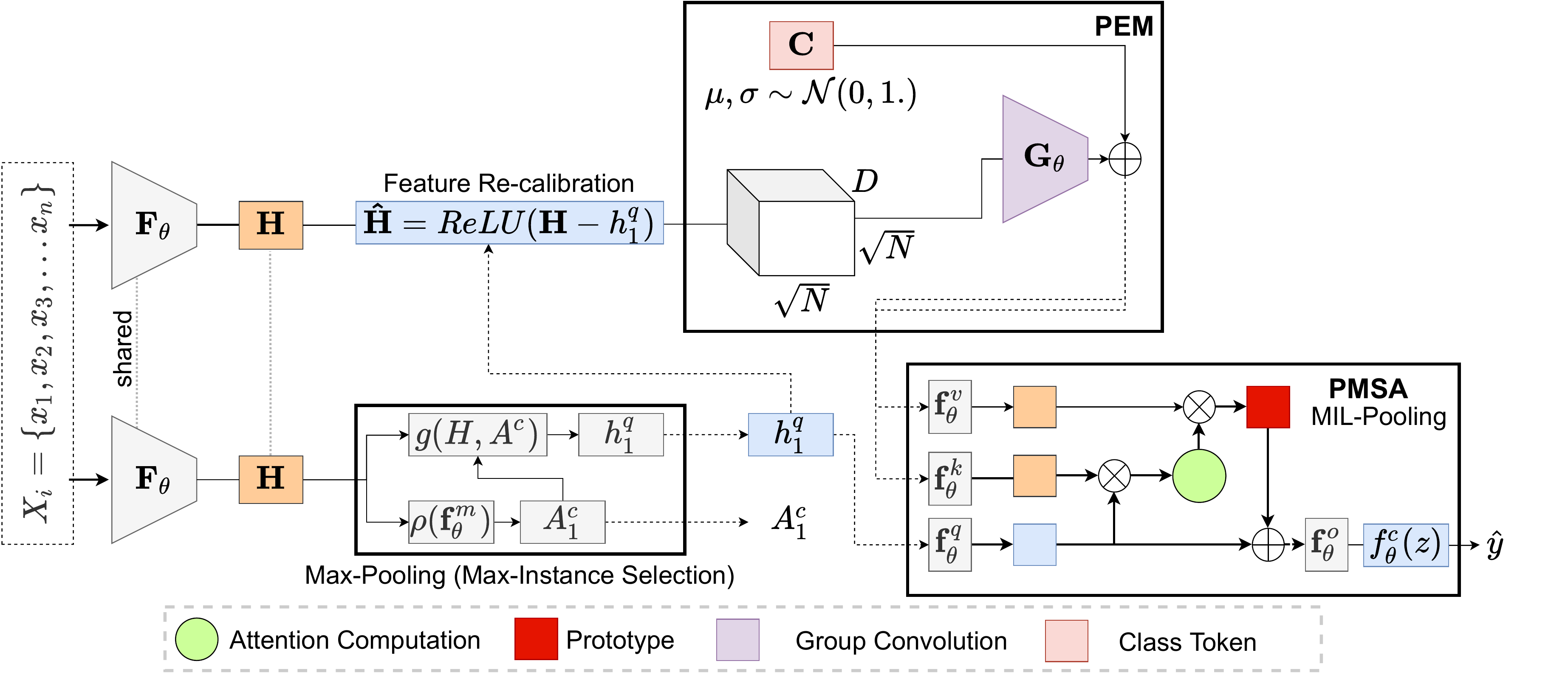}
	\caption{Overview of the proposed FR-MIL framework.} 
	\label{fig1}
\end{figure}

\noindent{\textbf{Overview}}. In this work, we consider a set of WSIs $\mathbf{X} = \{X_i\}$, each associated with a slide-level label $Y_i = \{0,1\}$, and our goal predict the slide labels using MIL (see. Fig. \ref{fig1}). We first extract instance features $\mathbf{H} \in \mathbb{R}^D$ using a neural network $\bf{F}_{\theta}$ i.e., $\bf{H}_i = \bf{F}_{\theta}(X_i)$, where $\bf{F}$ is either pre-trained on ImageNet or self-supervised learning\cite{chen2020simple,grill2020bootstrap}. In FR-MIL, we feed $\mathbf{H}$ to our max-instance selection module to obtain the highest instance (critical) as well as it's probability, then we re-calibrate the features $\mathbf{H}$ with the max-instance to obtain $\mathbf{\hat{H}}$. The position encoding module (PEM) creates a spatial representation of $\mathbf{\hat{H}}$, applies a single group convolution $\mathbf{G}_{\theta}$ to obtain correlated features, and then concatenates with a learnable class token $\mathbf{C}$. Finally, we perform MIL Pooling by Multi-head Self-Attention (PSMA) using the max-instance as a query and the output of $\mathbf{G}_{\theta}$ as key-value pairs to obtain the bag feature. FR-MIL is trained to minimize the bag loss, max-instance loss, and feature magnitude loss between positive and negative instance features. We detail each step below.     
   
\noindent{\textbf{Preliminaries: A Simple Baseline}}. Inspired by the work of Lee \textit{et al.}\cite{lee2021weakly} that employ feature magnitudes for uncertainity estimation of background actions in video sequences, we hypothesize that normal and positive bags should have different magnitudes and can serve as a simple MIL baseline. Herein, given instance features $\mathbf{H}^{c}_{i} = \{h_1,h_2,\dots,h_n\}$, where $c$ denotes the WSI class; the mean feature magnitude per WSI can be obtained as $\mu^{c}_{i} = \frac{1}{N}  \sum||\mathbf{H}^c_{i}||^2_{2}$, where $N$ denotes the number of instances in a bag. To obtain the probability $\mathbf{P}(y = 1.)$ of a bag, we formalize our assumption as:

\begin{equation}
	\mathbf{P}(y = 1.|\mu^{c}_{i}) = \frac{\text{min}(\tau,\mu^{c}_{i})}{\tau}, \label{eq:1}
\end{equation} 

\noindent where $\tau$ is the pre-defined maximum feature magnitude determined on the train-set only i.e., point at which the distributions first meet (see Fig.\ref{fig2}). Also, Eq. \ref{eq:1} is ensured to fall between 0 and 1, i.e., $0 \le \mathbf{P}(y = 1.|\cdot) \le 1$. In Fig. \ref{fig2}, we show the magnitudes on a given train-set (Camelyon16\cite{bejnordi2017diagnostic}) as a density plot. Note that while both normal and tumor slide curves appear to follow the Gaussian distribution (Fig. \ref{fig2}(a)), separation is non-trivial due to the presence of more normal than tumor instances ($\le10\%$), and reports a low accuracy ($44\%$) with $\tau=18.8$. In Fig. \ref{fig2}(b), we show that re-calibrating the distribution by subtracting the max-instance feature before computing the magnitudes creates better separability. Formally, $\hat{\mu}^{c}_{i} = \frac{1}{N}  \sum||\mathbf{\hat{H}}^c_{i}||^2_{2}$, where $\mathbf{\hat{H}}^{c}_{i} = \{\mathbf{\hat{H}} - h^c_{\text{max}}\}$ given $h^c_{\text{max}} = \text{argmax}_{c}~  \mathbf{\hat{H}}^c_{i}$. Notably, re-calibration improves the test-accuracy by $+39$ with $\tau=8.2$. Finally, Fig. \ref{fig2}(c) shows the learned distribution of FR-MIL when trained with a feature magnitude loss $\mathcal{L}_{fm}$ and re-calibration, with more significant improvements ($+44$), further validating our hypothesis. 

\noindent{\textbf{Feature Re-calibration \& Max-Instance Selection}}. Given the set of instance features $\mathbf{H}$, our goal is to select the max-instance $h^q$ and it's associated score $A^c$ using an instance classifier $\mathbf{f}^m_{\theta}$ in our Max-Pooling module (see. Fig. \ref{fig1}). Here, $A^c = \rho(\mathbf{f}^m_{\theta}(\mathbf{H}))$ where $\rho$ denotes the sigmoid function. Consequently, the sorted scores are used to index the max-instance in $\mathbf{H}$ via an operator $g(\cdot)$, with $h^q$ later employed for feature re-calibration, as well as instance feature aggregation via PSMA for bag prediction. The max score $A^c$ is used to train the instance classifier $\mathbf{f}^m_{\theta}$ using the loss $\mathcal{L}_{max}$, in parallel with other modules in FR-MIL. Formally, re-calibration of features can be modeled as 

\begin{equation}
	\mathbf{\hat{H}} = \text{ReLU}(\mathbf{\hat{H}} - h^q), \label{eq:2}
\end{equation}  

\noindent similar to the intuition highlighted by the simple baseline. To further incorporate the concept of distribution re-calibration in our framework, we draw connections to prior work\cite{lee2021weakly} for anomaly detection i.e., assumes the feature magnitudes of positive- and normal-bags are different, and can be modeled via uncertainty. Therefore, to effectively model the ambiguous normal/background features, the training procedure should employ both positive and negative bags simultaneously instead of selecting normal features within a single bag (single batch). Herein, counter to existing methods that use a single bag for training, we employ a sampling strategy to produce balanced bags per epoch i.e., we initialize a zero-tensor with the maximum bag size in during training, and fill the relevant bag instance features. Note that by `\textit{balanced}', we imply 1-negative and 1-positive bag is sampled. Formally, to enforce feature discrepancy we propose feature magnitude loss $\mathcal{L}_{fm}$ as: 

\begin{equation}
	\mathcal{L}_{fm}(\mathbf{\hat{H}}^{pos}_i,\mathbf{\hat{H}}^{neg}_i,\tau) = \frac{1}{N} \sum^{N}_{n=1}(\text{max}(0,\tau - ||\mathbf{\hat{H}}^{pos}_i||) + ||\mathbf{\hat{H}}^{neg}_i|| ), \label{eq:5}
\end{equation}

\noindent where $\mathbf{\hat{H}}^{pos}$, and $\mathbf{\hat{H}}^{neg}$ are the positive- and negative bag instance features, and $\tau$ is the pre-defined margin, respectively. While prior work\cite{li2021dual} equally used max-pooling to select the max-instance, note that non-local masked attention was proposed for bag feature learning, whereas we use PSMA and propose feature re-calibration. 

\noindent{\textbf{Positional Encoding Module (PEM)}}. In the standard transformer\cite{vaswani2017attention,dosovitskiy2020image} design, encoding spatial information has proved useful for recognition tasks. However, it is non-trivial for WSIs due to varying sizes. In this work, we employ a conditional position encoder (PEM)\cite{li2021dt} that takes re-calibrated features $\mathbf{\hat{H}}$, performs zero-padding to provide absolute position for a convolution $\mathbf{G}_{\theta}$, and later concatenates the output with a class token $\mathbf{C}$ initialized from the normal distribution. Here, features $\mathbf{\hat{H}}$ are re-shaped into a 2D image by first computing $\{H,W\}$ i.e., $H = \sqrt{N} = \sqrt{n}$, where $n$ is the number of instances in a bag. Thus, 

\begin{equation}
	\mathbf{\hat{H}} \in \mathbb{R}^D \rightarrow \mathbf{\hat{H}} \in \mathbb{R}^{B\times C\times H\times W}, \label{eq:3}
\end{equation} 
 
\noindent where $B$ is the batch-size, $C=D$ are the instance feature dimensions, and $\mathbf{G}_{\theta}$ is 2D convolution layer that performs  group convolution with kernel size $3 \times 3$, and $1 \times 1$ zero padding. Note that prior work\cite{li2021dt} used different sized convolutions in a pyramidal fashion. Instead, we opt for a single layer to maintain computational feasibility. Finally, let $\mathbf{\acute{H}} = \text{concat}(\mathbf{C}_{\theta},\mathbf{G}_{\theta}(\mathbf{\hat{H}}))$, where $\mathbf{\acute{H}} \in \mathbb{R}^{(N+1) \times D}$ are the flattened restored features i.e., in the case of a single bag.        

\noindent{\textbf{MIL Pooling by Multi-head Self-Attention (PMSA)}}. In order to pool instance features $\mathbf{\acute{H}}$ to a single bag feature, we employ a single multi-head Transformer encoder\cite{lee2019set} that takes as input the max-instance feature $h^q$ as a query and $\mathbf{\acute{H}}$ as key-value pairs i.e., $\text{PMSA}_{\theta}(h^q,\mathbf{\acute{H}})$. The formulation proposed by Lee \textit{et al.}\cite{lee2019set} for set-based tasks employs an attention function $\varphi(\mathbf{Q},\mathbf{K},\mathbf{V})$ to measure similarity between a query vector $\mathbf{Q}$ with key-value pairs $\mathbf{K,V} \in \mathbb{R}^{d\times m}$ as: $\varphi(\mathbf{Q},\mathbf{K},\mathbf{V}) = \text{softmax}(\frac{\mathbf{Q}\mathbf{K}^T}{\sqrt{m}})\mathbf{V}$, where $\{d,m\}$ is the instance feature dimension. This can be easily extended to multi-head attention by first projecting vectors onto $k$ different dimensions. The encoder consists of feed-forward networks $\{\mathbf{f}^q_{\theta},\mathbf{f}^k_{\theta},\mathbf{f}^v_{\theta}\}$, where $\mathbf{f}^o_{\theta}$ is fed the output of $\varphi$ (prototype) together with residual connections and optional Layer Normalization\cite{ba2016layer} (LN). Formally, let $\hat{\varphi} = \varphi(\mathbf{Q},\mathbf{K},\mathbf{V}) + \mathbf{Q}$, then:

\begin{equation}
	z = \text{PSMA}(h^q,\mathbf{\acute{H}},\mathbf{\acute{H}}) = \text{LN}(\hat{\varphi} + \text{ReLU}(\mathbf{f}^{o}_{\theta}(\hat{\varphi}))), \label{eq:4}
\end{equation}

\noindent to produce a bag feature $z$, later fed to the bag classifier $\mathbf{f}^c_{\theta}$ for WSI classification. Finally, FR-MIL is trained to minimize the bag-, max-pooling and feature losses. Thus, the final objective is:
 
\begin{equation}
\mathcal{L} = \gamma_1 \mathcal{L}_{bag}(\hat{y},y) + \gamma_2 \mathcal{L}_{max}(A^{c},y) + \gamma_3 L_{fm}(\mathbf{\hat{H}}^{pos}_i,\mathbf{\hat{H}}^{neg}_i
,\tau), 
\end{equation}
	
\noindent where $\{\gamma_i\}$ are balancing weights and $\mathcal{L}_{\{bag,max\}}$ is the binary cross-entropy loss over the true WSI labels $y$ given $\hat{y} = \mathbf{f}^c_{\theta}(z)$, respectively.

\section{Experiments}

\noindent{\textbf{Datasets}}. To demonstrate the effectiveness of FR-MIL, we conducted experiments on the publicly available dataset CAMELYON16\cite{bejnordi2017diagnostic}, and an in-house curated dataset termed COLON-MSI i.e., colorectal (adenocarcinoma) cancer slides involving microsatellite instable (MSI) molecular phenotypes\cite{boland2010microsatellite}. CAMELYON16 dataset was proposed for metastatis detection in breast cancer, it consists of 271 training sets and 129 testing sets. After pre-processing, a total of 3.2 million patches at $\times$20 magnification, with an average of 8,800 patches per bag, and a maximum of 30,000 patches per bag on the training set. On the other hand, COLON-MSI consists of both microsatellite-stable (MSS) and microsatellite-instable (MSI), and is thus a subtyping task. It consists of 625 images, split as follows: 360 training, 92 validation, and 173 testing sets. Experts pathologists detected the presence of tumors with Immunohistochemical analysis (IHC) and PCR-based amplification and collectively agreed on the final slide-level label. Note that tumor ROIs are not used in this work. After pre-proessing, a total of 3.5 million patches at $\times$20 magnification, an average of 6,000 patches/bag, and a maximum of 8900 patches in the train-set.   

\noindent{\textbf{Implementation Settings}}. In the pre-processing step, we extracted valid patches of $256\times256$ after tissue detection and discard patches with $\le15\%$ tissue entropy. For the instance encoder $\mathbf{F}_{\theta}$, we employed the SimCLR\cite{chen2020simple} ResNet18\cite{he2016deep} encoder trained by Lee \textit{et al.}\cite{li2021dual} for the CAMELYON16 dataset. On the COLON-MSI set, we used an ImageNet pre-trained ResNet18. Thus, each instance feature is represented as $\mathbf{H}_i \in \mathbb{R}^{n \times 512}$. FR-MIL is trained with balanced batch sampling (B = 2), and learning rate of $1e-4$ with Adam optimizer for 100 epochs with $20\%$ dropout as regularization, and PSMA has heads $k=8$. Hyper-parameters $\{\gamma_{1,2,3}\} = 0.33$, with $\tau = 8.48$ for CAMELYON16, and $\tau = 57.5$ on COLON-MSI, respectively.   

\noindent{\textbf{Comparison Methods}}. We compare FR-MIL to traditional MIL methods max- and mean-pooling\cite{wang2018revisiting}, as well as existing state-of-the-art methods: ABMIL\cite{ilse2018attention}, DSMIL\cite{li2021dual}, CLAM-SB\cite{lu2021data}, MIL-RNN\cite{campanella2019clinical}, and TransMIL\cite{shao2021transmil}. All compared methods are trained for 200 epochs on COLON-MSI with similar settings. 

\begin{table}[t]
	\centering
	\caption{
		Evaluation of the proposed method on CAMELYON16 (CM16) and COLON-MSI sets. Metrics accuracy (ACC) and area under the curve (AUC) were employed. $\dagger$ :\footnotesize{\emph{denotes scores reported in the paper using ResNet50 as $\mathbf{F}_{\theta}$}} with ImageNet features. 
	}
	\label{tab:1}
	\begin{tabular}{l cccc ccccc}
		\toprule
		& \multicolumn{4}{c}{CM16} && \multicolumn{4}{c}{COLON-MSI} \\
		\cmidrule{3-10} 
		Method && ACC && AUC &&& ACC && AUC  \\
		\midrule
		Mean-pooling\cite{wang2018revisiting} && 0.7984 && 0.7620 &&& 0.624 && 0.830  \\
		Max-pooling\cite{wang2018revisiting}  && 0.8295 && 0.8641 &&& 0.763 && 0.859  \\
		ABMIL\cite{ilse2018attention}        && 0.8450 && 0.8653 &&& 0.740 && 0.779  \\
		MIL-RNN\cite{campanella2019clinical}      && 0.8062 && 0.8064 &&& 0.630 && 0.631  \\
		CLAM-SB\cite{lu2021data}      && 0.845  && 0.894  &&& 0.786 && 0.820  \\
		DSMIL\cite{li2021dual}        && 0.8682 && 0.8944 &&& 0.734 && 0.811  \\
		TransMIL\cite{li2021dt}     && 0.791  && 0.813  &&& 0.676 && 0.617  \\
		TransMIL$\dagger$\cite{li2021dt}  && 0.8837  && 0.9309  &&& - && -  \\
		\midrule
		FR-MIL (w/ $\mathcal{L}_{bag}$) && 0.8600  && 0.8990  &&& 0.809 && 0.880  \\
		FR-MIL (w/ $\mathcal{L}_{bag} + \mathcal{L}_{fm}$) && 0.8760  && 0.8990  &&& 0.775 && 0.842  \\
		FR-MIL (w/ $\mathcal{L}_{bag} + \mathcal{L}_{max}$) && 0.8840  && 0.8940  &&& 0.780 && 0.831  \\
		\midrule
		FR-MIL (w/ $\mathcal{L}_{bag} + \mathcal{L}_{max} + \mathcal{L}_{fm}$) && 0.8910  && 0.8950  &&& 0.809 && 0.901  \\
		\bottomrule
	\end{tabular}
\end{table}

\section{Results and Discussion}

\noindent{\textbf{Main Results}}. Table \ref{tab:1} presents the results of our approach against recent methods. On the CAMELYON16 dataset, FR-MIL reports $+3\%$ (ACC) improvement over DSMIL with comparable performance ($+1\%$) to the reported TransMIL scores using a larger feature extractor. Given that only a small portion of a each positive bag contains tumors, using the max-instance for bag pooling with re-calibration is intuitively sound and shows better performance over other methods. Moreover, though we employ PEM similar to TransMIL, the use of a single PEM module on calibrated facilitates better correlation learning. 

\begin{figure}[t!]
	\begin{tabular}{ccc}
		\includegraphics[width=0.32\linewidth]{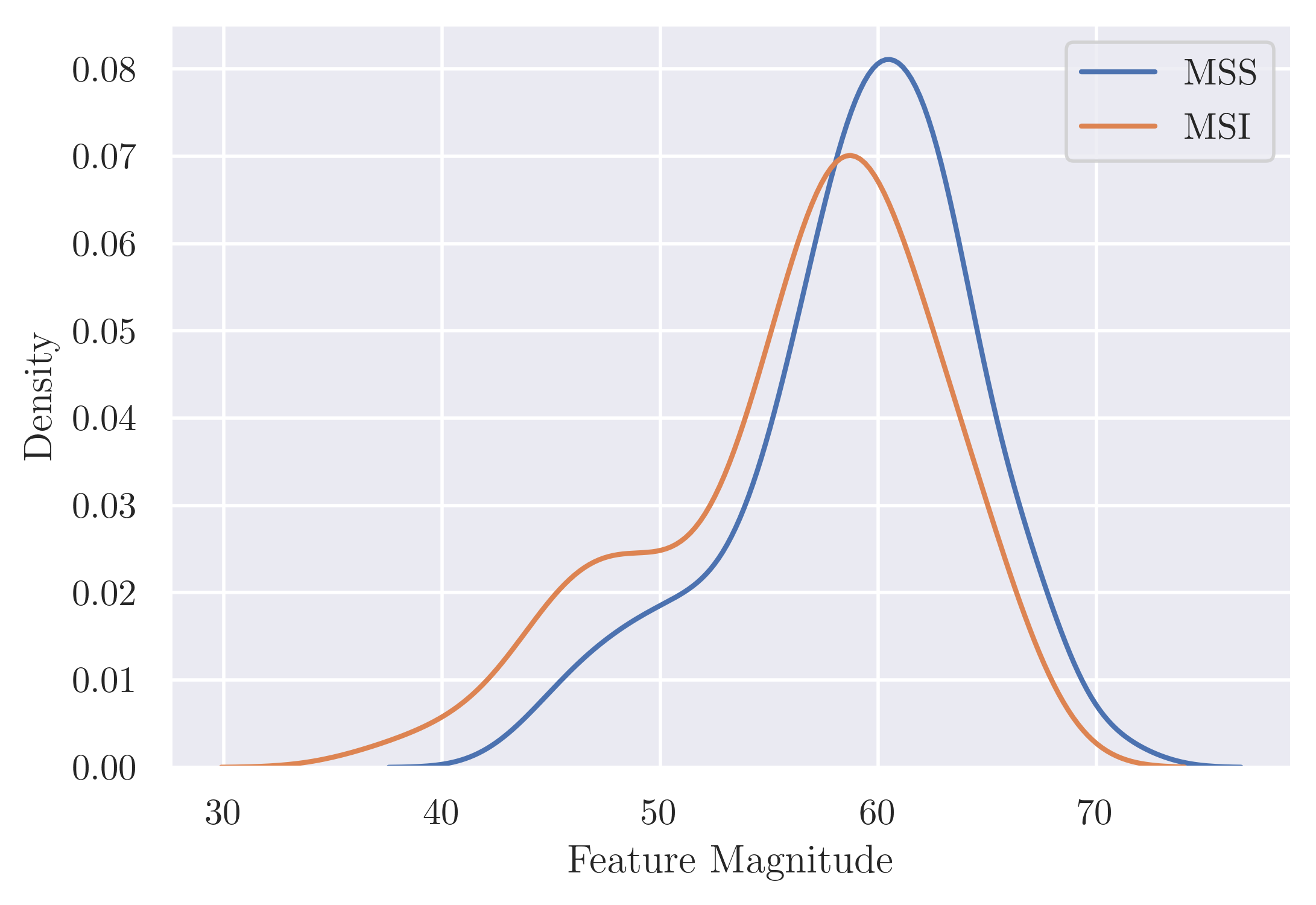} &
		\includegraphics[width=0.32\linewidth]{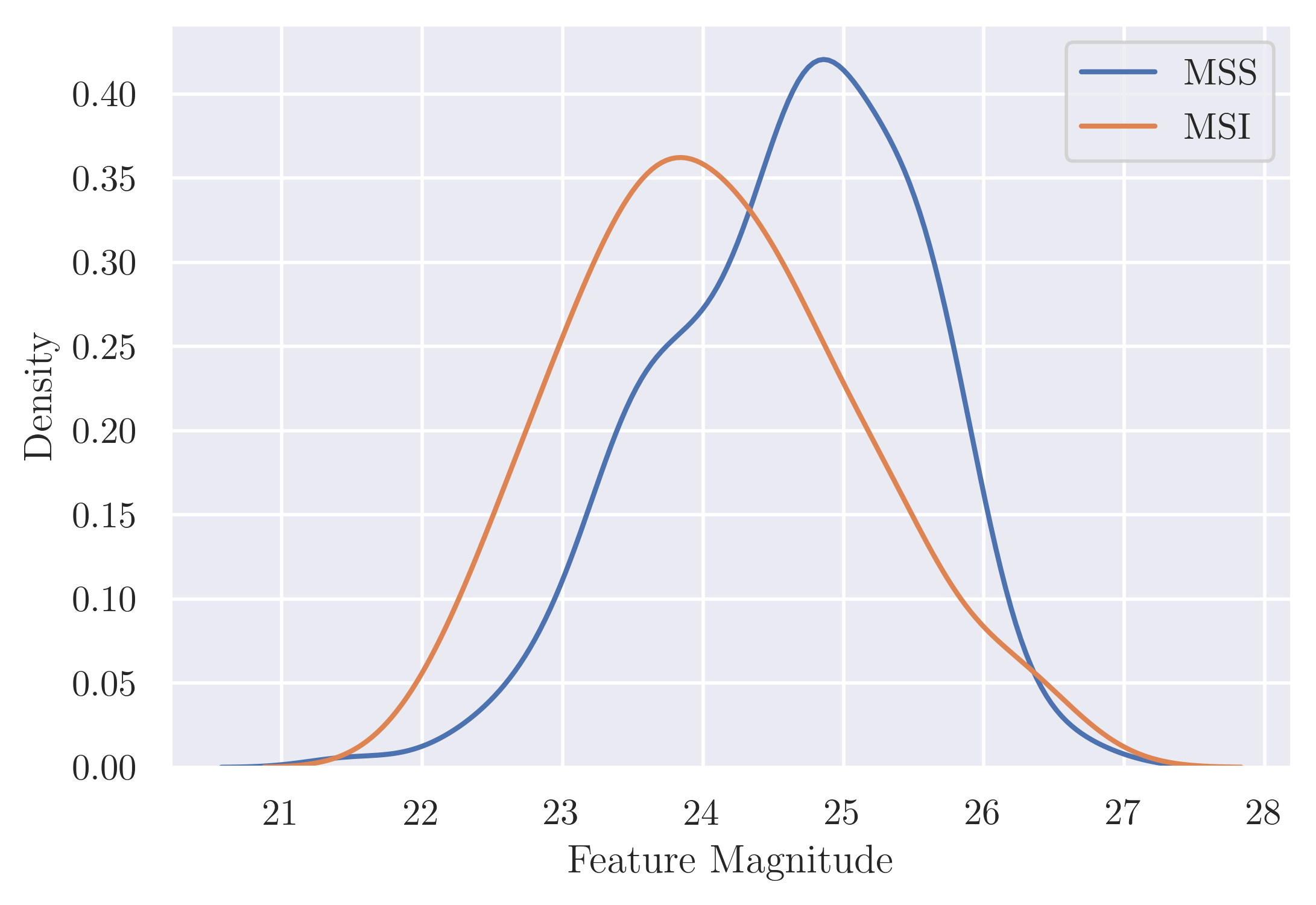} &
		\includegraphics[width=0.32\linewidth]{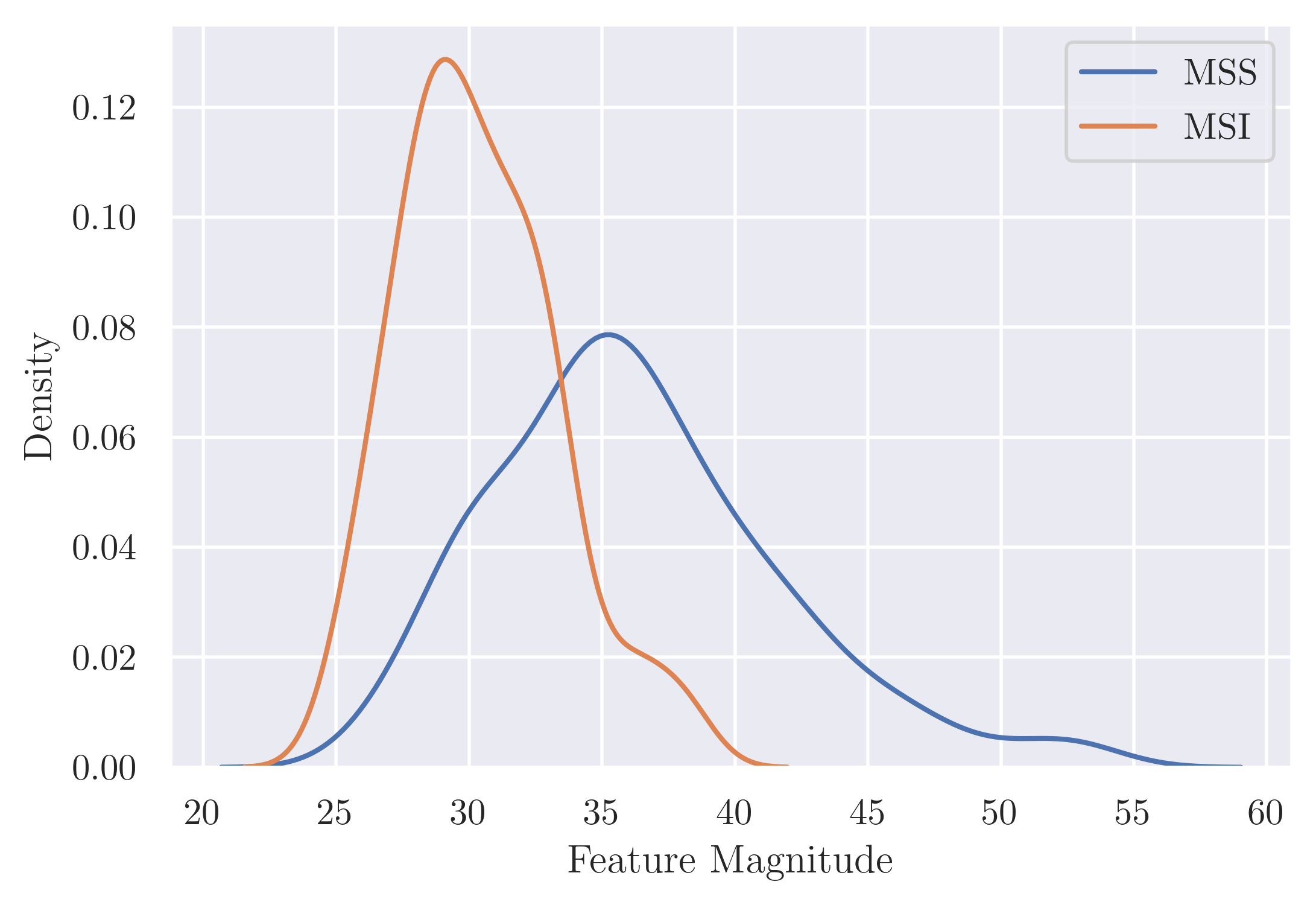} 
		\\
		(a) & (b) & (c)
	\end{tabular}
	\caption{Density plots of the mean feature magnitudes on the COLON-MSI train-set. (a) Original feature magnitudes. (b)  Max-instance re-calibration based features. (c) Features learned by our FR-MIL model.} \label{fig3}
\end{figure}

On the other hand, since the majority of slides contain relatively large tumor regions (averagely $\ge 75\%$) in COLON-MSI, max- and mean-pooling show high AUC but had inconsistent results (ACC). Overall, CLAM-SB reports the best ACC among the compared methods i.e., $78.6\%$. Interestingly, TransMIL performed poorly on this set, possibly due to over-parametrization, and the morphological similarities between MSS and MSI instances. Similar observations were drawn regarding MIL-RNN. Consequently, the proposed FR-MIL highlights the importance of calibration in subtyping problems, reporting $+2\%$ (ACC) and $+5\%$ (AUC) improvements achieving the best scores. See Fig. \ref{fig3} for the distribution of features.

\noindent{\textbf{Ablations}}. To validate the effectiveness of the proposed losses on learning, we evaluated FR-MIL with/without certain losses (see. Table \ref{tab:1}). First, on CAMELYON16, we found $\mathcal{L}_{fm}$ was a crucial component to further boost the AUC score. Overall, when both $\mathcal{L}_{max}$ and $\mathcal{L}_{fm}$ were omitted, performance drops were noted. On the other hand, on COLON-MSI, using a $\mathcal{L}_{bag}$ only had the best scores, whereas the use of both  $\mathcal{L}_{max}$ and $\mathcal{L}_{fm}$ resulted in a significant reduction (AUC). However, employing all the losses resulted in more stable scores.  

\section{Conclusion}

In this work, we presented a MIL framework for Whole Slide Image classification that leverages Feature Re-calibration, applicable to both binary and sub-typing tasks. We show that: (i) by leveraging feature magnitude discrepancy between positive and negative bags as a probabilistic measure; a simple baseline is comparable in performance to classic MIL operators, (ii) explicitly re-calibrating the data distribution with max-instances during training by drawing connections to the standard MIL assumption is simple yet effective, and (iii) the use of a metric feature loss to encourage better feature separation in (+/-) bags improves both Accuracy and AUC over state-of-the-art methods. Further exploring the utility of this approach in multi-scale setting, or designing an adaptable margin (mean magnitude) estimator will be topics of future research.
\\
\\
\noindent{\textbf{Acknowledgments}} This work was supported by the DGIST R\&D program of the Ministry of Science and ICT of KOREA (21-DPIC-08), Smart HealthCare Program funded by the Korean National Police Agency (220222M01), and IITP grant funded by the Korean government (MSIT) (No.2021-0-02068, Artificial Intelligence Innovation Hub).

\bibliographystyle{splncs04}
\bibliography{macros,refer}

\begin{thebibliography}{10}
\providecommand{\url}[1]{\texttt{#1}}
\providecommand{\urlprefix}{URL }
\providecommand{\doi}[1]{https://doi.org/#1}

\bibitem{amores2013multiple}
Amores, J.: Multiple instance classification: Review, taxonomy and comparative
  study. Artificial intelligence  \textbf{201},  81--105 (2013)

\bibitem{ba2016layer}
Ba, J.L., Kiros, J.R., Hinton, G.E.: Layer normalization. arXiv preprint
  arXiv:1607.06450  (2016)

\bibitem{banerji2022deep}
Banerji, S., Mitra, S.: Deep learning in histopathology: A review. Wiley
  Interdisciplinary Reviews: Data Mining and Knowledge Discovery
  \textbf{12}(1),  e1439 (2022)

\bibitem{bejnordi2017diagnostic}
Bejnordi, B.E., Veta, M., Van~Diest, P.J., Van~Ginneken, B., Karssemeijer, N.,
  Litjens, G., Van Der~Laak, J.A., Hermsen, M., Manson, Q.F., Balkenhol, M.,
  et~al.: Diagnostic assessment of deep learning algorithms for detection of
  lymph node metastases in women with breast cancer. Jama  \textbf{318}(22),
  2199--2210 (2017)

\bibitem{boland2010microsatellite}
Boland, C.R., Goel, A.: Microsatellite instability in colorectal cancer.
  Gastroenterology  \textbf{138}(6),  2073--2087 (2010)

\bibitem{campanella2019clinical}
Campanella, G., Hanna, M.G., Geneslaw, L., Miraflor, A., Werneck Krauss~Silva,
  V., Busam, K.J., Brogi, E., Reuter, V.E., Klimstra, D.S., Fuchs, T.J.:
  Clinical-grade computational pathology using weakly supervised deep learning
  on whole slide images. Nature medicine  \textbf{25}(8),  1301--1309 (2019)

\bibitem{chalapathy2019deep}
Chalapathy, R., Chawla, S.: Deep learning for anomaly detection: A survey.
  arXiv preprint arXiv:1901.03407  (2019)

\bibitem{chen2021pixel}
Chen, H., Wang, K., Zhu, Y., Yan, J., Ji, Y., Li, J., Xie, D., Huang, J.,
  Cheng, S., Yao, J.: From pixel to whole slide: Automatic detection of
  microvascular invasion in hepatocellular carcinoma on histopathological image
  via cascaded networks. In: MICCAI. pp. 196--205. Springer (2021)

\bibitem{chen2020simple}
Chen, T., Kornblith, S., Norouzi, M., Hinton, G.: A simple framework for
  contrastive learning of visual representations. In: ICML. pp. 1597--1607.
  PMLR (2020)

\bibitem{chikontwe2020multiple}
Chikontwe, P., Kim, M., Nam, S.J., Go, H., Park, S.H.: Multiple instance
  learning with center embeddings for histopathology classification. In:
  MICCAI. pp. 519--528. Springer (2020)

\bibitem{dimitriou2019deep}
Dimitriou, N., Arandjelovi{\'c}, O., Caie, P.D.: Deep learning for whole slide
  image analysis: an overview. Frontiers in medicine p.~264 (2019)

\bibitem{dosovitskiy2020image}
Dosovitskiy, A., Beyer, L., Kolesnikov, A., Weissenborn, D., Zhai, X.,
  Unterthiner, T., Dehghani, M., Minderer, M., Heigold, G., Gelly, S.,
  Uszkoreit, J., Houlsby, N.: An image is worth 16x16 words: Transformers for
  image recognition at scale. In: ICLR (2021)

\bibitem{fan2021learning}
Fan, L., Sowmya, A., Meijering, E., Song, Y.: Learning visual features by
  colorization for slide-consistent survival prediction from whole slide
  images. In: MICCAI. pp. 592--601. Springer (2021)

\bibitem{feng2021mist}
Feng, J.C., Hong, F.T., Zheng, W.S.: Mist: Multiple instance self-training
  framework for video anomaly detection. In: CVPR. pp. 14009--14018 (2021)

\bibitem{grill2020bootstrap}
Grill, J.B., Strub, F., Altch{\'e}, F., Tallec, C., Richemond, P., Buchatskaya,
  E., Doersch, C., Avila~Pires, B., Guo, Z., Gheshlaghi~Azar, M., et~al.:
  Bootstrap your own latent-a new approach to self-supervised learning. NeurIPS
   \textbf{33},  21271--21284 (2020)

\bibitem{he2016deep}
He, K., Zhang, X., Ren, S., Sun, J.: Deep residual learning for image
  recognition. In: CVPR. pp. 770--778 (2016)

\bibitem{he2012histology}
He, L., Long, L.R., Antani, S., Thoma, G.R.: Histology image analysis for
  carcinoma detection and grading. Computer methods and programs in biomedicine
   \textbf{107}(3),  538--556 (2012)

\bibitem{ilse2018attention}
Ilse, M., Tomczak, J., Welling, M.: Attention-based deep multiple instance
  learning. In: ICML. pp. 2127--2136. PMLR (2018)

\bibitem{lee2019set}
Lee, J., Lee, Y., Kim, J., Kosiorek, A., Choi, S., Teh, Y.W.: Set transformer:
  A framework for attention-based permutation-invariant neural networks. In:
  ICML. pp. 3744--3753. PMLR (2019)

\bibitem{lee2021weakly}
Lee, P., Wang, J., Lu, Y., Byun, H.: Weakly-supervised temporal action
  localization by uncertainty modeling. In: AAAI. vol.~2 (2021)

\bibitem{li2021dual}
Li, B., Li, Y., Eliceiri, K.W.: Dual-stream multiple instance learning network
  for whole slide image classification with self-supervised contrastive
  learning. In: CVPR. pp. 14318--14328 (2021)

\bibitem{li2021comprehensive}
Li, C., Li, X., Rahaman, M., Li, X., Sun, H., Zhang, H., Zhang, Y., Li, X., Wu,
  J., Yao, Y., et~al.: A comprehensive review of computer-aided whole-slide
  image analysis: from datasets to feature extraction, segmentation,
  classification, and detection approaches. arXiv preprint arXiv:2102.10553
  (2021)

\bibitem{li2021dt}
Li, H., Yang, F., Zhao, Y., Xing, X., Zhang, J., Gao, M., Huang, J., Wang, L.,
  Yao, J.: Dt-mil: Deformable transformer for multi-instance learning on
  histopathological image. In: MICCAI. pp. 206--216. Springer (2021)

\bibitem{lu2021data}
Lu, M.Y., Williamson, D.F., Chen, T.Y., Chen, R.J., Barbieri, M., Mahmood, F.:
  Data-efficient and weakly supervised computational pathology on whole-slide
  images. Nature biomedical engineering  \textbf{5}(6),  555--570 (2021)

\bibitem{rymarczyk2021kernel}
Rymarczyk, D., Borowa, A., Tabor, J., Zielinski, B.: Kernel self-attention for
  weakly-supervised image classification using deep multiple instance learning.
  In: IEEE Winter. Conf. Application. Comput. Vis. pp. 1721--1730 (2021)

\bibitem{shao2021transmil}
Shao, Z., Bian, H., Chen, Y., Wang, Y., Zhang, J., Ji, X., et~al.: Transmil:
  Transformer based correlated multiple instance learning for whole slide image
  classification. NeurIPS  \textbf{34} (2021)

\bibitem{sharma2021cluster}
Sharma, Y., Shrivastava, A., Ehsan, L., Moskaluk, C.A., Syed, S., Brown, D.:
  Cluster-to-conquer: A framework for end-to-end multi-instance learning for
  whole slide image classification. In: Medical Imaging with Deep Learning. pp.
  682--698. PMLR (2021)

\bibitem{shi2020loss}
Shi, X., Xing, F., Xie, Y., Zhang, Z., Cui, L., Yang, L.: Loss-based attention
  for deep multiple instance learning. In: AAAI. vol.~34, pp. 5742--5749 (2020)

\bibitem{srinidhi2021deep}
Srinidhi, C.L., Ciga, O., Martel, A.L.: Deep neural network models for
  computational histopathology: A survey. Medical Image Analysis  \textbf{67},
  101813 (2021)

\bibitem{vaswani2017attention}
Vaswani, A., Shazeer, N., Parmar, N., Uszkoreit, J., Jones, L., Gomez, A.N.,
  Kaiser, {\L}., Polosukhin, I.: Attention is all you need. NeurIPS
  \textbf{30} (2017)

\bibitem{wang2018revisiting}
Wang, X., Yan, Y., Tang, P., Bai, X., Liu, W.: Revisiting multiple instance
  neural networks. Pattern Recognition  \textbf{74},  15--24 (2018)

\bibitem{yang2020free}
Yang, S., Liu, L., Xu, M.: Free lunch for few-shot learning: Distribution
  calibration. In: ICLR (2020)

\end{thebibliography}

\end{document}